\documentclass[11pt,a4paper]{article}
\usepackage[printwatermark]{xwatermark}
\usepackage{amsmath,amssymb}
\usepackage{booktabs}
\usepackage[hyperref]{eacl2021}
\usepackage{latexsym}
\usepackage{hyperref}
\usepackage{multirow,multicol}
\usepackage{times}
\usepackage{pifont}
\usepackage{xcolor}
\usepackage{xspace}
\usepackage{csquotes}

\usepackage{textcomp}
\usepackage{fontawesome}
\usepackage{stfloats}

\DeclareFontFamily{U}{skulls}{}
\DeclareFontShape{U}{skulls}{m}{n}{ <-> skull }{}
\newcommand\skull{{\usefont{U}{skulls}{m}{n}\symbol{'101}}}

\aclfinalcopy

\usepackage{todonotes}
\newcounter{maarten}

\newcommand{\draftonly}[1]{#1}
\renewcommand{\draftonly}[1]{}

\usepackage{transparent}
\usepackage{soul}

\newcommand{\profanities}{swear words\xspace}

\newcommand{\noi}{\textsc{nOI}\xspace}
\newcommand{\oni}{\textsc{OnI}\xspace}
\newcommand{\oi}{\textsc{OI}\xspace}
\newcommand{\toxt}{\textsc{ToxTrig}\xspace}

\newcommand{\aae}{\textsc{aae}\xspace}
\newcommand{\wae}{\textsc{wae}\xspace}

\newcommand{\learnedmixin}{\textsc{Learned-Mixin}\xspace}
\newcommand{\lmixin}{\textsc{LMixin}\xspace}

\title{
Challenges in Automated Debiasing for Toxic Language Detection 
}

\author{
Xuhui Zhou$^{\heartsuit}$\quad 
Maarten Sap$^{\clubsuit}$\quad 
Swabha Swayamdipta$^{\diamondsuit}$\quad 
Noah A. Smith$^{\clubsuit}$$^{\diamondsuit}$\quad
Yejin Choi$^{\clubsuit}$$^{\diamondsuit}$\\
\\
$^{\heartsuit}$Department of Linguistics, University of Washington \\
  $^{\clubsuit}$Paul G.\ Allen School of Computer Science \& Engineering, University of Washington \\
   $^{\diamondsuit}$Allen Institute for Artificial Intelligence\\
   {\tt xuhuizh@uw.edu,\{msap,yejin,nasmith\}@cs.washington.edu,  swabhas@allenai.org } 
}

\begin{document}

\maketitle

\begin{abstract}
 \textit{\textbf{Warning}: this paper contains content that may
 be offensive or upsetting.}

 Biased associations have been a challenge in the development of classifiers for detecting toxic language, hindering both fairness and accuracy.  
 As potential solutions, we investigate recently introduced debiasing methods for text classification datasets and models, as applied to toxic language detection.
 Our focus is on lexical (e.g., swear words, slurs, identity mentions) and dialectal 
 markers (specifically African American English).
 Our comprehensive experiments establish that existing methods are limited in their ability to prevent biased behavior in current toxicity detectors.
We then propose an automatic, dialect-aware data correction method, as a proof-of-concept study. 
Despite the use of synthetic labels, this method reduces dialectal associations with toxicity.
Overall, our findings show that debiasing a model trained on biased toxic language data is not as effective as simply relabeling the data to remove existing biases.
\end{abstract}

\section{Introduction}
\label{sec:intro}

\begin{figure}[t]
\centering
\includegraphics[width=\columnwidth]{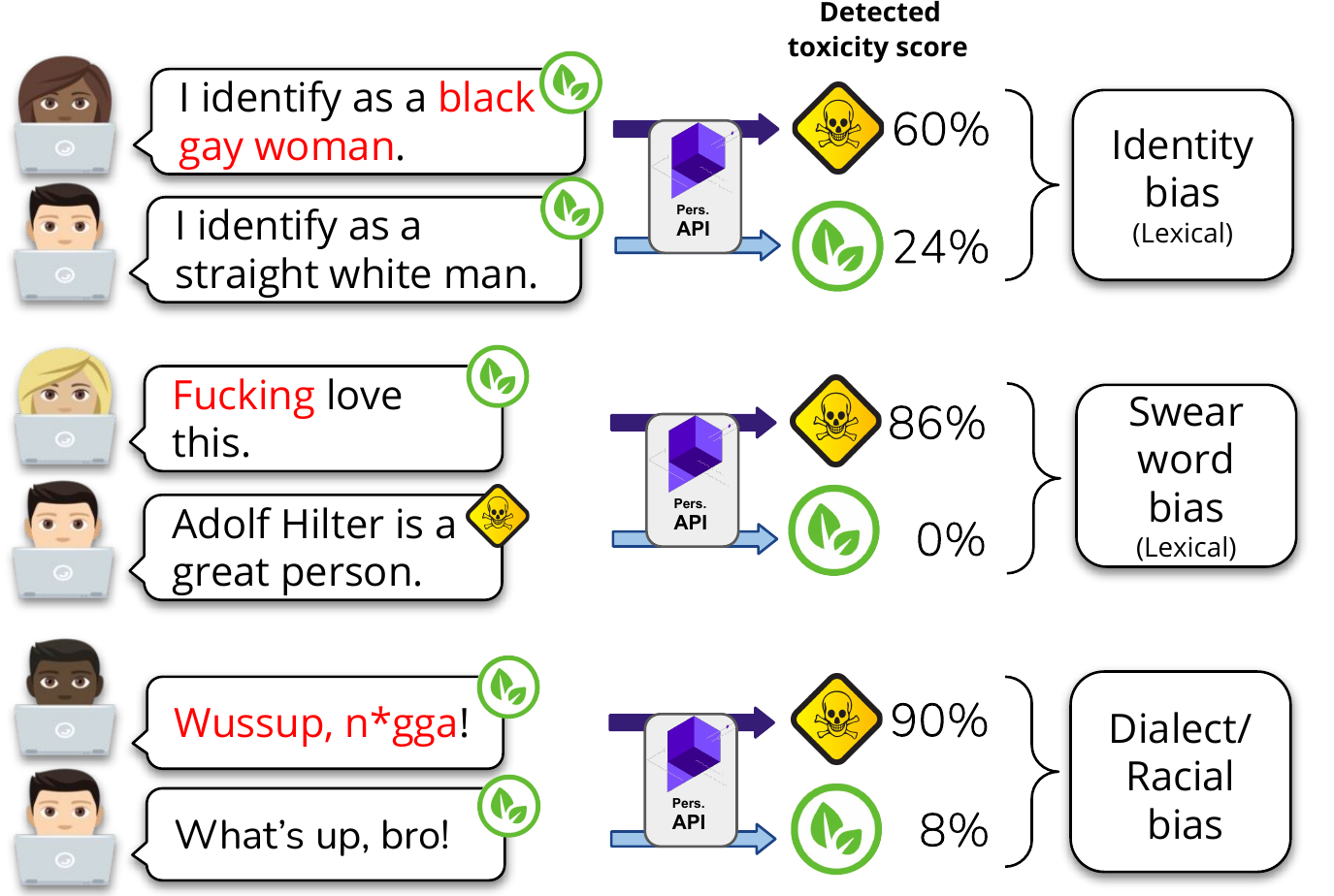} 
\caption{
Lexical items and dialect markers cause problematic behavior for toxic language detection systems such as the widely used \href{https://www.perspectiveapi.com/}{PerspectiveAPI}.
In the top two example pairs, statements with minority identity mentions and swear words used inoffensively are flagged as toxic, but majority identity mentions or offensive statements without overt swearing are missed.
The bottom pair shows dialect-based racial bias for two inoffensive greetings, where markers of African American English (\aae) trigger the toxicity detector. 
}
\label{fig:introFig}
\end{figure}

Current hate speech or toxic language detection\footnote{We use \emph{hate speech} and \emph{toxic language} interchangeably in this work, though their definitions do not perfectly align.} systems exhibit problematic and discriminatory behavior that causes them to have disparate negative impact on minority populations \cite{Yasin2018black,guynn2020facebookBoycott, Kim2020IntersectionalBI, DiasOliva2020}.
Tweets simply containing a minority identity mention are commonly flagged as toxic by current systems, in contrast to those containing majority identity mentions, as illustrated in Figure \ref{fig:introFig}.

At the core of the issue are \textit{dataset biases}, i.e., spurious correlations between surface patterns and annotated toxicity labels (\S\ref{sec:datasets_biases}), which stem from the data creation process \cite{sap2019risk}.
Previous work has outlined two such biases for hate speech datasets (both shown in Figure \ref{fig:introFig}): \textit{lexical bias} which associates toxicity with the presence of certain words \cite[e.g., profanities, identity mentions;][]{Dixon2018MeasuringAM,dinan-etal-2019-build} and \textit{dialectal bias}, where toxicity is correlated with surface markers of African American English \citep[\aae;][]{davidson-etal-2019-racial,sap2019risk}.
When trained on biased datasets, models acquire \textit{and} exacerbate these biases \citep[e.g., flagging text by Black authors as more toxic than by white authors;][]{sap2019risk,zhang-2018-mitigating}.

Concurrently, there has been elevated interest in developing \textit{debiasing methods} for standard natural language understanding (NLU) tasks, i.e., methods that aim to decrease over-reliance on spurious correlations in NLU models \cite{clark-etal-2019-dont,he-etal-2019-unlearn,karimi-mahabadi-etal-2020-end,bras2020}.
This raises a natural question: \textit{are current debiasing approaches effective for mitigating biases specific to toxic language detection?}

In this work, we address the above question by investigating two classes of debiasing approaches to mitigate lexical and dialectal biases---one that employs additional training objectives for bias removal, and another that filters training instances likely exhibiting spurious biases (\S\ref{sec:debiasing methods}).
Through comprehensive experiments, we show that both approaches face major challenges in mitigating biases from a model trained on a biased dataset (in our case, the dataset from \citealp{founta2018}) for toxic language detection.
While data filtering results in reduced bias associations in the data, \textit{models} trained on filtered datasets still pick up on lexical (\S\ref{sec:lexical_experiments}) and dialectal biases (\S\ref{sec:experiments_dialectal}).
We find that dialectal biases are particularly challenging to address, as has also been shown by \citet{xia-etal-2020-demoting}. 
``Debiased'' models still disproportionately flag text in certain dialects as toxic. 
Notably, mitigating dialectal bias through current debiasing methods does not mitigate a model's propensity to label tweets by Black authors as more toxic than by white authors.

We additionally explore an alternative proof-of-concept study---relabeling supposedly toxic training instances whose automatic translations into a majority dialect are deemed non-toxic by the classifier.
To this end, we create a synthetic dataset via few-shot dialect translation system built with GPT-3 \cite{brown2020language}.
While only an illustrative solution, it nevertheless takes into account the dialectal context of the tweet, resulting in a model less prone to dialectal and racial biases (\S\ref{sec:gpt-3}).
Overall, our findings indicate that debiasing a model already trained on biased toxic language data can be challenging, compared to relabeling the data to remove existing biases.
Our code and data are publicly available on Github.\footnote{\url{https://github.com/XuhuiZhou/Toxic_Debias}}

\section{Biases in Toxic Language Detection}
\label{sec:datasets_biases}

We test the use of debiasing\footnote{Our definition of ``bias'' is specific to the social biases in toxic language detection datasets, grounded as lexical and dialectal biases; see \citet{Blodgett2020-oy} for a detailed investigation of the term ``bias''.} methods for the task of toxic language detection, which aims to flag rude, offensive, hateful, or toxic language on the internet, with the goal of moderating online communities \cite{roberts2019behind,vidgen2019much}.
This task differs in several ways from the natural language understanding (NLU) tasks that debiasing methods have been successful on, such as textual entailment \cite[e.g., SNLI, MNLI;][]{bowman-etal-2015-large,williams-etal-2018-broad} or reading comprehension \cite[e.g., SQuAD;][]{rajpurkar2016squad}.
First, compared to these NLU tasks where there is one correct label, the toxicity of language is inherently more nuanced, subjective, and contextual, which causes toxic language datasets to have lower agreement in general \cite{Ross2017measuring}.
Second, the dataset biases in NLU are predominantly artifacts introduced during data creation \cite[e.g., negations, exaggerations;][]{schwartz2017effect,Gururangan2018AnnotationAI}, whereas those in toxic language detection are grounded in the social dynamics of the world \cite{Spears1998africanamerican,Technau2018-ld}.
For example, viewing \aae as a more toxic or less proper variety of English is a form of linguistic discrimination that upholds racial hierarchies in the United States \citep{Rosa2017-ec}.

In this work, we consider two broad categories of toxic language dataset biases---lexical (\S\ref{ssec:lexical-bias}) and dialectal (\S\ref{ssec:dialect-bias}).
Our experiments focus on a single, widely used dataset (\S\ref{ssec:data}) from \citet{founta2018}.

\subsection{Lexical Biases (\toxt)}
\label{ssec:lexical-bias}
Current toxic language detection systems often rely on the presence or absence of certain words (e.g., swear words, identity mentions) to make their predictions \cite{Dixon2018MeasuringAM,dinan-etal-2019-build}. %
While most previous analyses of this bias relied on a simple list of ``bad'' words \cite{davidson-etal-2019-racial,dinan-etal-2019-build},\footnote{\url{https://tinyurl.com/list-of-bad-words}} 
we take a more nuanced view of how lexical items can convey toxicity, inspired by work in pragmatics and sociolinguistics of rudeness \cite[\textit{inter alia}]{Dynel2015-wn,Kasper1990-lb}.
Specifically, we manually split our full list of words into three distinct categories depending on the extent to which they carry profane or hateful meanings or are simply associated with hateful contexts.\footnote{We note, however, that this categorization is in itself subjective.}
We refer to the full set of words as \toxt, for Toxicity Triggers, which is included in our released repository.\footnote{\url{https://github.com/XuhuiZhou/Toxic_Debias/blob/master/data/word_based_bias_list.csv}}

\paragraph{Non-offensive minority identity mentions (\noi)} refers to descriptive mentions of minoritized demographic or social identities (e.g., \textit{gay}, \textit{female}, \textit{Muslim}). 
While these mentions are not usually inherently offensive by themselves, they are often found in offensive statements that are hateful towards minorities \citep{Dixon2018MeasuringAM}. 
We detect these identity mentions in text using a list of 26 regular expressions.

\paragraph{Possibly offensive minority identity mentions (\oi)} 
are mentions of minoritized identities that could denote profanity or hate depending on pragmatic and contextual interpretations.
This includes slurs and objectifying outdated terms to refer to minority groups, which are usually understood as attacks.
Additionally, this includes \textit{reclaimed} slurs (\textit{queer}, \textit{n*gga}), which connote less offensive intent 
when spoken by in-group members compared to out-group members \cite{Croom2013-uc}.

\paragraph{Possibly offensive non-identity mentions (\oni)} 
contains \profanities and other profanities, which are usually offensive but not associated to any social groups (e.g., \textit{f*ck}, \textit{sh*t}).
Note that the pragmatic interpretation of these words is not necessarily always toxic or offensive \cite{Dynel2012Swearing}, as they are often used to convey closeness between the speaker and listener or emphasize the emotionality of a statement (e.g., second example in in Figure \ref{fig:introFig}).

\subsection{Dialectal Biases (\aae)}
\label{ssec:dialect-bias}
Current toxic language detection systems also associate higher toxicity with dialectal markers of African American English \citep[\aae;][]{sap2019risk,davidson-etal-2019-racial}.
Since \aae is a variety of English that is common among African Americans and often signals a cultural identity in the US  \cite{Green2002aae}, this dialect-based racial bias causes speech by Black authors to be suppressed more often than non-Black authors \cite{sap2019risk}, thereby exacerbating racial inequality \cite{rosa2019looking}.

In our experiments, we estimate the dialect of a tweet using a topic model from \citet{blodgett-etal-2016-demographic}. 
This model was trained on 60M tweets, where the dialect of the tweet was inferred from the model coordinates, which yielded a probability of a tweet being in one of four dialects (African-American English, white-aligned English, Hispanic, and other).
In this study, we only focus on African-American English (\aae) and white-aligned English (\wae) tweets; both definitions are based on US English, as per \citet{blodgett-etal-2016-demographic}.\footnote{We avoid using disputed terms such as \emph{general American English}, \emph{standard American English}, or \emph{mainstream US English}, which are frequently used for \wae, since we believe that no dialect should be privileged with the designation ``general'', ``standard", or ``mainstream" \cite{rosa2019looking}.%
}
Our experiments either use the probability of a tweet being in these dialects, or assign tweets their estimated-most-probable dialect.

\subsection{Dataset for Toxic Language Detection}
\label{ssec:data}
We focus our analyses on a widely used hate speech dataset of English tweets \cite{founta2018}.
The tweets were collected using a multi-round bootstrapping procedure, and were labeled out of context\footnote{Only the tweet text---no profile information or conversational context---was shown to annotators.}
for toxic language. 
We focus on the 86k tweets that are annotated as hateful, abusive, or neither and discard those labelled as spam.
We aggregate the abusive and hateful labels into a single \textit{toxic} category, yielding 32k toxic and 54k non-toxic tweets.\footnote{We also explored using another widely used hate speech dataset \cite{davidson2017}, which collected tweets using a seed list of swear words and slurs. However, in line with findings by \citet{xia-etal-2020-demoting}, debiasing led to degenerate behavior due to the data collection process, as discussed in Appendix \ref{supp:davidson-data}.}

\section{Debiasing Methods}
\label{sec:debiasing methods}
We consider two types of debiasing methods from current literature.
The first type addresses known, pre-defined biases---such as lexical and dialectal biases for hate speech detection, via a model-based approach involving additional training objectives (\S\ref{sec:predefined}).
In contrast, the second type is agnostic to prior knowledge about biases, and instead filters out examples that appear ``too easy'' and might hence contain spurious correlations (\S\ref{sec:unspecified}).

\subsection{Debiased Training for Pre-Defined Toxicity Biases}
\label{sec:predefined}
We use the \learnedmixin method of \citet{clark-etal-2019-dont}, which achieved high out-of-distribution (OOD) performance on several NLU tasks, for debiased training.
This method trains an ensemble containing a \textit{bias-only} model which only uses pre-defined features corresponding to known biases, and a \textit{full} model which uses all features.
Intuitively, the ensemble encourages the full model to rely more on features unrelated to the biases.
Once trained, the bias-only model is discarded, and only the ``bias-free'' full model is used for inference, following \citet{clark-etal-2019-dont}.

\paragraph{Bias-only model} 
Given its effectiveness on bag-of-words (BoW) features, we use an SVM classifier as the lexical-bias-only model.
For example, the \toxt-only model counts the frequency of \toxt words in each tweet.
Our dialectal-bias-only model uses the probability of dialects (\aae, \wae, Hispanic, and other) obtained from a dialect detector \cite{blodgett-etal-2016-demographic} as features in a SVM classifier. 

\paragraph{Full model} 
We fine-tune a RoBERTa-large classifier \cite{liu2019roberta}, a state-of-the-art classifier for the toxicity detection task. 
See Appendix \ref{supp:predefined} for more modeling details.

Note that %
we only consider the \learnedmixin-\oni and \learnedmixin-\toxt models for lexical debiasing, due to poor accuracies of the bias-only models for \noi and \oi.\footnote{The \noi and \oi bias-only models reach 63\% and 67\% accuracy, respectively, which is empirically hard for the ensemble to use.  %
This is likely due to low coverage in the train set of those categories (4.43\% \noi and 4.25\% \oi).}

\subsection{Data Filtering for Spurious Biases}
\label{sec:unspecified}

In addition to debiasing methods that handle known biases, we also explore automated approaches which filter out instances exhibiting unspecified, spurious biases.
Specifically, we describe below two data selection methods that have shown strong OOD performance.

\paragraph{AFLite \citep{bras2020}} 
is an algorithm based on the key intuition that examples predicted correctly by the simplest methods likely exhibit spurious biases. %
An ensemble of simple linear classifiers is trained and tested on different partitions of the data; test instances which are ``predictable'', or classified correctly by most classifiers in the ensemble are discarded.
The algorithm is iterative, and is repeated until a target data size is achieved.
Models trained on this filtered dataset achieve higher performance on OOD and adversarially constructed test sets, compared to the original model, on several text and image classification datasets.
This indicates a reduction in spurious biases in the filtered data.

\paragraph{DataMaps \cite{swayamdipta2020datamaps}} show the presence of distinct regions in a dataset---namely, easy, hard and ambiguous---defined with respect to a given model.
These regions are discovered based on the training dynamics of a model, determined by the model's confidence in the true class, for each example, as well as the variability of this confidence, throughout training epochs.
\citet{swayamdipta2020datamaps} show that training exclusively on the hard and ambiguous regions of the data results in high OOD performance, indicating lower prevalance of spurious biases.
The easy region is the largest in size for RoBERTa; however, experiments showed that training exclusively on these examples hurt OOD generalization on different NLU tasks.
Following this work, we create DataMaps-Easy, DataMaps-Ambiguous, and DataMaps-Hard subsets for our dataset. 

Following \citet{swayamdipta2020datamaps}, we set the target filtered subset size to 33\% of the original training set for both filtering methods, but our filtering additionally preserved the original label proportions.
We then fine-tune a RoBERTa-large classifer on these filtered subsets; see Appendix \ref{supp:unspecified} for more details.

\section{Experiments: Lexical Biases}
\label{sec:lexical_experiments}

We investigate the effect of debiasing approaches (\S\ref{sec:debiasing methods}) on removing lexical biases in hate speech detection.
First, we discuss the evaluation framework for measuring bias reduction (\S\ref{sec:evaluation}).
We present quantitative (\S\ref{sec:results_lexical_bias}) and qualitative (\S\ref{sec:lexical_qualitative}) results on lexical bias removal for all debiasing approaches, and OOD evaluation for debiased training methods (\S\ref{sec:adv_evaluation}).
See Appendix \ref{supp:training_setting} 
for hyperparameters and other experimental settings.

\subsection{Evaluation Framework}
\label{sec:evaluation}

We report the performance of all models as overall accuracy and $F_1$ with respect to the toxic class.
Given that current hate speech systems tend to rely heavily on the presence of \noi, \oi, and \oni mentions (\S\ref{ssec:lexical-bias}) for labeling text as toxic, we use false positive rate (FPR) over each of these categories to measure the degree of bias in the model, following \citet{hardt2016equal} and \citet{xia-etal-2020-demoting}.
Specifically, we report the FPR of a model on tweets containing \noi (FPR$_\noi$), \oi (FPR$_\oi$), and \oni (FPR$_\oni$), as well the $F_1$ corresponding to each of these classes.
Intuitively, the lower the FPR$_{*}$, the less the model infers lexical associations for toxicity, and hence is less biased.

\begin{table}
\centering
\small
\begin{tabular}{clccc}
\toprule
& & $R_{\noi} \downarrow$ & $R_{\oi} \downarrow$ & $R_{\oni} \downarrow$\\
\cmidrule{3-5}
& Original  & 0.0445 & 0.2641 & 0.6718 \\
\midrule[0.03em]
\multirow{5}{*}{\rotatebox{90}{{33\% train}}} & Random  & 0.0345 & 0.2603 & 0.6683\\
& AFLite  & 0.0434 & 0.2458 & 0.6016\\
& DataMaps-Ambig. & 0.0126 & 0.1968 & \textbf{0.5839}\\
& DataMaps-Hard  & \bf{0.0081} & \textbf{0.1853} & 0.5849\\
& DataMaps-Easy & 0.0772 & 0.3661 & 0.7720\\
\bottomrule
\end{tabular}
\caption{\label{tab:intrin_eval_word}
Lexical associations between toxicity and \toxt mentions in the original dataset \cite{founta2018} and various filtered counterparts. 
Random, AFLite, and DataMaps all contain only 33\% of the original data after filtering. 
Lower Pearson $R$ correlation value indicates less superficial patterns in the dataset, i.e., less bias.
\textbf{\textit{Takeaway:}} The hard and ambiguous subsets given by DataMaps contain the lowest amount of lexical associations, indicated in boldface.
}
\end{table}

\begin{table*}[t]
\centering
\small
\begin{tabular}{clcccrccccc}
\toprule
& & \multicolumn{2}{c}{Test (12893)} & \multicolumn{2}{c}{\noi (602)} & \multicolumn{2}{c}{\oi (553)} & \multicolumn{2}{c}{\oni (3236)}\\
\cmidrule(lr){3-4} \cmidrule(lr){5-6} \cmidrule(lr){7-8} \cmidrule(lr){9-10}
& & Acc.$\uparrow$ & $F_1\uparrow$ & $F_1\uparrow$ & FPR$_{\noi} \downarrow $ & $F_1\uparrow$ & FPR$_{\oi} \downarrow$ & $F_1 \uparrow$ & FPR$_{\oni} \downarrow$  \\
\midrule
& Vanilla & 94.21$_{0.0}$ & 92.33$_{0.0}$ & 89.76$_{0.3}$ & 10.24$_{1.3}$ & 98.84$_{0.1}$ & 85.71$_{0.0}$ & 97.34$_{0.1}$ & 64.72$_{0.8}$ \\
& \lmixin-\oni & 89.65$_{1.5}$ & 85.59$_{2.5}$ & 87.04$_{1.1}$ & 13.99$_{1.5}$ & 98.87$_{0.0}$ & 85.71$_{0.0}$ & 87.87$_{4.5}$ & \textbf{43.74}$_{3.1}$ \\
& \lmixin-\toxt & 90.44$_{0.7}$ & 86.94$_{1.1}$ & 85.47$_{0.3}$ & 11.15$_{1.7}$ & 97.64$_{0.3}$ & \textbf{71.43}$_{0.0}$ & 90.41$_{1.8}$ & 44.55$_{1.5}$ \\
\midrule[0.03em]
\multirow{5}{*}{\rotatebox{90}{{33\% train}}} 
& Random & 94.07$_{0.1}$ & 92.18$_{0.1}$ & 89.48$_{0.4}$ & 9.33$_{0.7}$ & \textbf{98.93}$_{0.0}$ & \textbf{83.33}$_{3.4}$ & 97.40$_{0.1}$ & 67.15$_{0.6}$ \\
& AFLite & 93.86$_{0.1}$ & 91.94$_{0.1}$ & \textbf{90.21}$_{0.4}$ & 11.26$_{1.1}$ & 98.90$_{0.0}$ & 85.71$_{0.0}$ & 97.32$_{0.1}$ & 67.97$_{3.4}$ \\
& DataMaps-Ambig. & 94.33$_{0.1}$ & 92.45$_{0.1}$ & 89.16$_{0.7}$ & 7.39$_{1.0}$ & 98.87$_{0.0}$ & 85.71$_{0.0}$ & \textbf{97.54}$_{0.0}$ & 64.39$_{1.4}$ \\
& DataMaps-Hard & \textbf{94.50}$_{0.0}$ &\textbf{ 92.61}$_{0.1}$ & 89.54$_{0.4}$ & {6.26}$_{0.9}$ & 98.84$_{0.0}$ & 85.71$_{0.0}$ & 97.43$_{0.0}$ & 61.95$_{1.1}$ \\
& DataMaps-Easy & 94.00$_{0.1}$ & 91.94$_{0.2}$ & 86.81$_{0.6}$ & \textbf{5.92}$_{0.7}$ & 98.87$_{0.0}$ & \textbf{83.33}$_{3.4}$ & 97.17$_{0.1}$ & \textbf{60.33}$_{3.8}$ \\
\bottomrule

\end{tabular}
\caption{\label{tab:results_lexical}
Evaluation of lexical bias removal for all debiasing methods on the \citet{founta2018} test set.
Results show the mean and s.d. (subscript) of accuracy and $F_1$ across 3 runs, as well as $F_1$ and false positive rate exclusively on test examples containing specific \toxt mentions---\noi, \oi and \oni, along with the number of examples in each category.
The lower the FPR$_{\textbf{*}}$, the less the model infers lexical associations for toxicity.
The first block shows debiased training approaches, along with the vanilla classifier, which are trained on the full dataset.
The second block shows data filtering approaches, all trained on only 33\% of the training data.
Best performance in each block is boldfaced.
\textbf{\textit{Takeaway:}} While data filtering approaches achieve overall higher performance, debiased training approaches perform better on lexical bias reduction, in aggregate.
}
\end{table*}

\newcommand{\toxicIcon}{\skull\xspace}
\newcommand{\nonToxicIcon}{\faLeaf\xspace}

\begin{table*}[t]
\centering
\footnotesize
\begin{tabular}{@{}p{12cm}p{0.8cm}p{0.8cm}p{0.8cm}@{}}
\toprule
 & Gold & DM-Hard & DM-Easy\\
\midrule
@user THIS? LMAOOOOO...do yourself a solid and stay out of Black people's mentions and mind your caucasia... & \nonToxicIcon & \toxicIcon & \nonToxicIcon \\
RT @user I wish I wasn't so annoying like I even piss myself off & \toxicIcon & \nonToxicIcon  & \toxicIcon \\
@user If you want to attack people, attack fundamentalists of all faiths. Attack those who condemn 1.5bn people out of hand. & \nonToxicIcon & \toxicIcon & \nonToxicIcon  \\
\bottomrule
\end{tabular}
\caption{\label{tab:qualitative_lexical}
Examples of test set tweets with their gold-standard annotations and predictions from models trained on DataMaps-Hard (DM-Hard) and DataMaps-Easy (DM-Easy) subsets.
\toxicIcon denotes tweets with toxic labels, and \nonToxicIcon represents non-toxic labels. 
We anonymize the usernames to protect user privacy.
}
\end{table*}

\paragraph{Evaluation for Filtered Datasets}
\label{sec:eval_dataset}
We additionally consider metrics based on spurious lexical associations for data filtering approaches.
This measures prevalence of spurious surface patterns in the filtered datasets, which might propagate to models trained on the data. %
Specifically, we report the Pearson's   correlation between the 
gold standard toxicity label and whether or not it contains \noi, \oi, or \oni mentions. 
These correlations are denoted as $R_{\oni}$, $R_{\noi}$, and $R_{\oi}$, respectively; lower values indicate reduction in lexical biases.

\paragraph{Baselines}
\label{sec:baselines}
We consider comparison against two natural baselines: a vanilla RoBERTa-large classifier trained on the original dataset (Original). %
We also consider a baseline trained on a random selection of the training data (Random), for comparison with data filtering methods for debiasing.
Each subset is trained on 33\% of the training data.

\subsection{Results for Lexical Bias Reduction}
\label{sec:results_lexical_bias}

First, we measure the reduction in lexical biases in filtered datasets, as given by AFLite and DataMaps. 
As shown in Table \ref{tab:intrin_eval_word}, subsets given by AFLite and the ambiguous and hard regions produced by DataMaps reduce the overall associations between \toxt words and toxicity, compared to the original and random baselines; DataMaps-Hard has the largest reduction.
On the other hand, as expected, DataMaps-Easy shows an \emph{increased} association between \toxt mentions and toxicity, showing that the these %
examples display overt lexical biases.

Table \ref{tab:results_lexical} shows results for lexical bias reduction using both debiased training approaches, as well as models trained on datasets filtered using AFLite and all three regions from DataMaps.
Both debiased training approaches, \lmixin-\oni and \lmixin-\toxt, reduce FPR$_\oni$ as well as FPR$_\oi$ by a large amount. 
However, both approaches also hurt in-distribution test performance, indicating that \oni and other \toxt features are essential for good performance.\footnote{When we combine the bias-only model and the full model, we obtain competitive performance (see Appendix \ref{supp:wbias_only}).}
In contrast, the models trained on hard and ambiguous subsets from DataMaps both preserve in-distribution performance, even though they are trained only a third of the original data.
They also reduce the rate of falsely predicting \noi mentions as toxic (FPR$_\noi$), while not showing much improvement for \oni and maintaining FPR$_\oi$ of the original baseline.

Surprisingly, the model trained on the easy subset from DataMaps shows good bias reduction on the \noi and \oni categories, while matching the random selection baseline for \oi.
This is despite DataMaps-Easy showing an increased association between \toxt mentions and toxicity (Table \ref{tab:intrin_eval_word}).
Notably, the $F_1$ for all categories suffers under this model, indicating that it is less competent %
than the baseline.
These results suggest that reduced associations in the data might not necessarily lead to debiased models trained on the same data.
Overall, no single approach outperforms all others across different categories for lexical debiasing.

\subsection{Qualitative Analysis}
\label{sec:lexical_qualitative}

A qualitative study of the \citet{founta2018} test set shows the presence of many annotation errors. 
We show three representative annotation errors in Table \ref{tab:qualitative_lexical}.
The first example contains an atypical example of toxicity, towards white folks, which the annotators might have been unaware of. 
It also contains a link which annotators had access to, but not models.
The second contains the word \textit{p*ss} which the annotators may have relied for their assessment.
The third encourages violence/abuse towards an identity which isn't typically the target of violence.
Interestingly, the DataMaps-Easy predictions agree with all the gold standard annotations; perhaps such annotation errors and ambiguity are responsible for the performance discussed in \S\ref{sec:results_lexical_bias}.
These annotation ambiguities might also impair our measurement for models' performance and debiasing effects, and expose a limitation of these datasets.

\subsection{Adversarial Evaluation: \oni-Adv}
\label{sec:adv_evaluation}

\begin{figure}[t]
\centering
\includegraphics[width=\columnwidth]{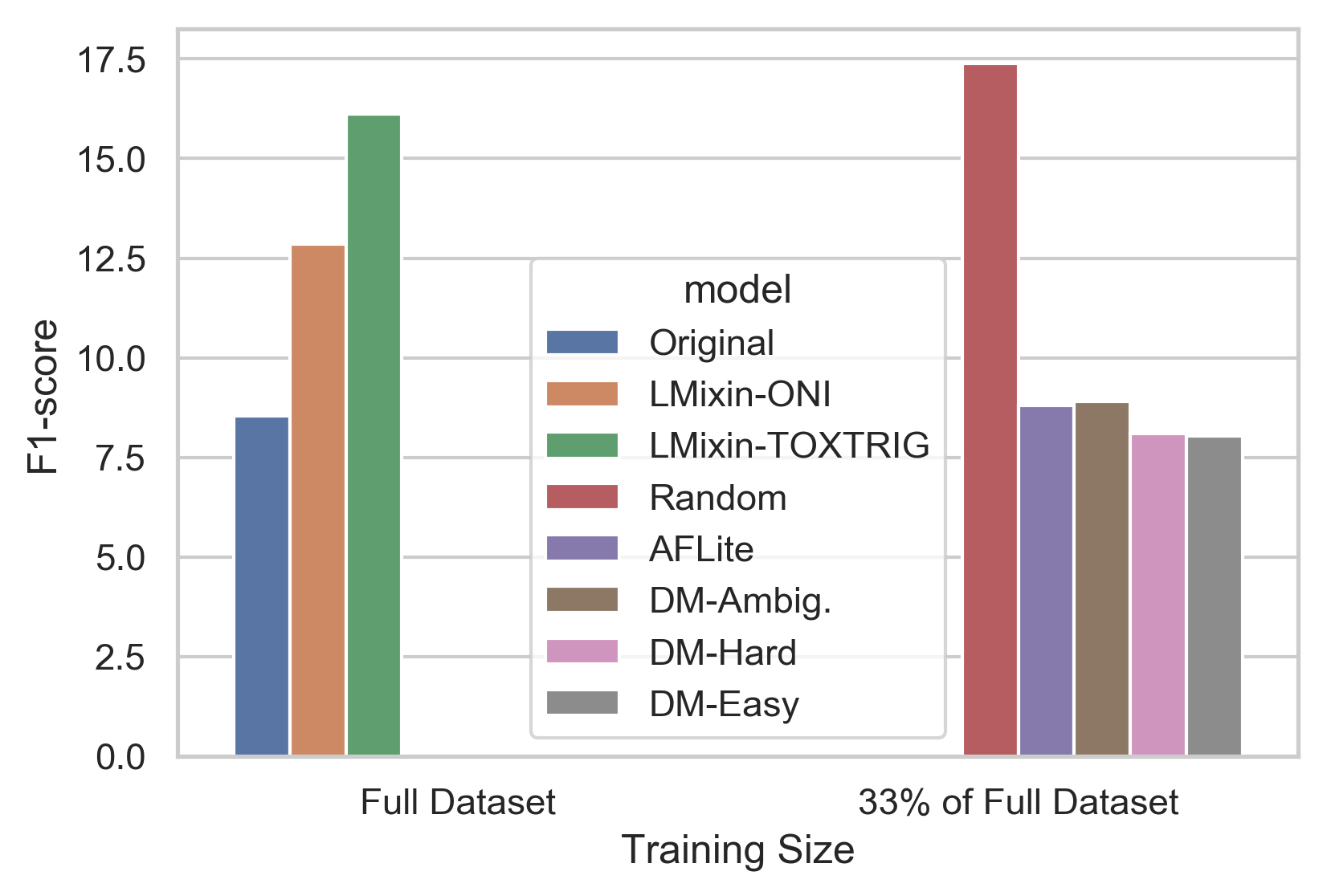} 
\caption{Challenge set evaluation for lexical biases, comparing all debiasing methods with baselines, using the \oni-Adv test set.
\textbf{\textit{Takeaway:}} $F_1 (\uparrow)$ measures show that all models perform poorly at identifying toxic text not containing overtly lexical cues of toxicity.
In general, debiased training approaches outperform the original model on this challenge set, while data filtering is not as effective.
}
\label{fig:Swear Adv}
\end{figure}

To further study the reliance of debiased models on the \oni words, we use the test set from \citet{dinan-etal-2019-build}, which contains 3000 manually created sentences, 300  toxic. 
The toxic sentences barely contain any \oni mentions,
e.g., \textit{Cheese is made by french people who smell}. %
We call this test set \oni-Adv (for adversarial) since it challenges models with a reversal in the association between toxicity and offensive non-identity words (e.g., ``\textit{f*ck}'', ``\textit{sh*t}''). 

We report $F_1$ for all models in Figure \ref{fig:Swear Adv}, which shows how well a model identifies toxicity in offensive tweets that do not contain overtly lexical cues of toxicity. %
The debiased training approaches improve over the baselines;  data filtering methods do not. 
One reason for this might be that data filtering methods were trained on much less data than both \lmixin models.
Regardless, none of the models we test are good at predicting subtle, non-overt toxicity.

\begin{table}[t]
\centering
\small
\begin{tabular}{@{}c@{\hspace{.7em}}lccc@{}}
\toprule
& &  & \multicolumn{2}{c}{Test} \\ %
\cmidrule(lr){4-5} %
& & $R_{\aae} \downarrow$  & $F_1 \uparrow$ & FPR$_\aae \downarrow$ \\ %
\midrule
& Vanilla & 0.4079 %
& 92.33$_{0.0}$ & 16.84$_{0.3}$ \\ %
& \lmixin-Dialect & - %
& 92.26$_{0.1}$ & {16.07}$_{0.4}$ \\ %
\midrule[0.03em]
\multirow{5}{*}{\rotatebox{90}{{33\% train}}} & Random & 0.4027 %
& 92.18$_{0.1}$ & 16.67$_{0.6}$ \\ %
& AFLite & 0.3577 %
& 91.94$_{0.1}$ & 16.84$_{0.8}$ \\ %
& DataMaps-Ambig. & 0.2965 %
& 92.45$_{0.1}$ & 15.99$_{0.4}$ \\ %
& DataMaps-Hard & \textbf{0.2878} %
& \textbf{92.61}$_{0.1}$ & \textbf{13.71}$_{0.2}$ \\ %
& DataMaps-Easy & 0.5347 %
& 91.94$_{0.2}$ & 19.46$_{2.8}$ \\ %
\midrule[0.03em]
& \aae-relabeled & 0.3453 %
& 91.64$_{0.3}$ & \textbf{12.69}$_{0.0}$ \\ %
\bottomrule
\end{tabular}
\caption{\label{tab:results_dialectal}
Dialectal bias evaluation for all debiasing methods (\S\ref{sec:experiments_dialectal}), as well as the relabeling approach (\S\ref{sec:gpt-3}) on the \citet{founta2018} test set.
We report $F_1$ and the false positive rate with respect to tweets in \aae (FPR$_{\aae}$), reflecting dialectal bias (lower is less biased), showing mean and s.d. (subscript) across 3 runs.
(Top Block) Debiased training approaches, along with the vanilla classifier, are all trained on the full dataset.
(Middle Block) Random, AFLite and DataMaps all are trained on only 33\% of the training data.
Best performance for each training set size is in boldface.
\textbf{\textit{Takeaway:}} Both debiasing approaches improve performance over baselines, with DataMaps-Hard proving the most effective at debiasing.
(Bottom Block) \aae-relabeling results in a model which despite following a noisy process yields even larger improvements for dialectal debiasing.
}
\end{table}

\section{Experiments: Dialectal and Racial Biases}
\label{sec:experiments_dialectal}
We test the efficacy of the bias reduction methods from \S\ref{sec:debiasing methods} for dialectal bias (\S\ref{ssec:dialect-bias}) reduction. %

\subsection{Dialectal Biases}
For our dialectal bias experiments, we first infer the dialect of a tweet as described in \S\ref{ssec:dialect-bias}.
Then, analogous to the lexical bias evaluation, we quantify the dialectal debiasing using the Pearson's correlation between estimated probabilities of \aae and toxicity ($R_{\aae}$), and the false positive rates of models on \aae tweets (FPR$_\aae$).
See Appendix \ref{supp:training_setting} for hyperparameters and other experimental settings.

Results in Table \ref{tab:results_dialectal} show that almost all data filtering and debiasing methods reduce dialectal biases, with DataMaps-Easy as the exception (consistent with Table \ref{tab:intrin_eval_word}).
Notably, DataMaps-Hard performs the best at dialectal debiasing, both in terms of toxicity-\aae correlation ($R_{\aae}$) and in terms of false flagging of toxicity (FPR$_\aae$).
Interestingly, most models' decrease in false flagging is small, suggesting room for improvement.

\begin{table}
\centering
\small
\begin{tabular}{@{}clccp{0.2cm}c@{}}
\toprule
& & {W-Tox.} & {AA-Tox.} & $\Delta\downarrow$ & {AA/W}$\downarrow$   \\
\cmidrule{3-6}
& Original       & 7.24         & 12.61     & 5.37     & 1.74         \\
& \lmixin-Dialect  & 7.50           & 12.55    & 5.06      & 1.67         \\
\midrule
\multirow{5}{*}{\rotatebox{90}{{33\% train}}} & Random         & 8.28          & 13.24    & 4.96      & 1.60      \\
& AFLite         & 7.32          & 11.64      & 4.33    & 1.59      \\
& DataMaps-Ambig.          & 6.75           & 12.17    & 5.42     & 1.80      \\
& DataMaps-Hard           & 6.36           & 11.67      & 5.31    & 1.84        \\
& DataMaps-Easy           & 8.46            & 16.30     & 7.83    & 1.94         \\
\midrule
& \aae-relabeled        & 6.93       & 10.60     & \textbf{3.67}     & \textbf{1.53}      \\
\bottomrule
\end{tabular}

\caption{\label{tab:userlevelrace18}
Racial disparity in toxicity prediction reported on \citet{preotiuc-pietro-ungar-2018-user}.
\textbf{W-Tox.} indicates \% of white users' tweets being flagged as toxic, \textbf{AA-Tox.} indicates \% of African American users' tweets being flagged as toxic, $\Delta$ refers to the difference between AA-Tox.~and W-Tox., and \textbf{AA/W} refers to the ratio between AA-Tox. and W-Tox.
\textbf{\textit{Takeaway:}} Methods generally fail in debiasing on this OOD test set except the relabeling approach shows some benefit.
}
\end{table}

\subsection{Racial Biases}
To quantify the real-world impact of dialect-based racial bias, we measure the rates of toxicity predicted by models on a corpus of tweets for which the race of authors is available, but not annotations of toxicity.
Specifically, we consider the dataset released by \citet{preotiuc-pietro-ungar-2018-user}, 
which consists of 5.4M tweets, collected from 4,132 survey participants (3,184 White, 374 African American) with self-reported race/ethnicity and Twitter user handles.\footnote{For efficiency, we randomly select 12k tweets from the dataset as the OOD test set.}

We quantify our models' racial bias by measuring the difference in rates of flagging tweets by African American authors and those by white authors, following \citet{sap2019risk}.\footnote{Note that we assume that authors from all races have the same likelihood of writing toxic language.}

Listed in Table \ref{tab:userlevelrace18}, our results show that automatic debiasing methods do not consistently decrease the racial discrepancy in flagging toxicity.
Notably, the toxicity rates on tweets by African American authors---and the diferences compared to white authors---are similar across all debiasing methods and baselines, except for DataMaps-Easy, which shows the most racial bias in toxicity flagging. %
Surprisingly, DataMaps-Hard, which mitigated dialectal bias the best out of all debiasing methods, also shows high discrepancy between author races.
Confirming previous results, this suggests that debiasing these systems requires more than automatic debiasing methods.

\section{Towards Data Relabeling}
\label{sec:gpt-3}

\begin{table*}[t]
\centering
\footnotesize
\begin{tabular}{p{6.5cm}p{6.5cm}p{0.4cm}p{0.4cm}}
\toprule
 \aae & GPT-3 \wae Translation & Gold & New \\
\midrule
RT @user I can't stand a bad texter bruh like don't be mad if I forget about yo ass & RT @user I can't stand a bad texter bro like don't be mad if I forget about you & \toxicIcon & \nonToxicIcon \\
RT @user Retweet if you fuck with this!!!! & RT @user Retweet if you like this! & \toxicIcon & \nonToxicIcon \\
RT @user That nigga needs anger management & RT @user That guy needs anger management & \toxicIcon & \nonToxicIcon  \\
RT @user oh fucking hell take a day off man & RT @user oh fuck take a day off man  & \toxicIcon & \toxicIcon\\
\bottomrule
\end{tabular}
\caption{\label{tab:qualitative_dialect}
Examples of \aae tweets with their GPT-3 based \wae translation, and original gold standard and new annotations based on \aae-relabeled.
For the first three tweets, the (biased) gold labels are changed by models predicting the new labels on their \wae translations. 
\toxicIcon\xspace indicates presence of toxicity, and \nonToxicIcon\xspace represents non-toxic. We anonymize the usernames to protect user privacy.
}
\end{table*}

Based on our quantitative and qualitative analyses, we believe there still is room for improvement in debiasing hate speech detection.  %
Therefore, we turn our attention to the role of label noise in datasets. 
Partly inspired by our qualitative analyses of debiased models' predictions, we design a proof-of-concept study where we automatically correct the label of tweets using a(n automatic) dialectal translation of the tweet, inspired by previous work showing that highlighting \aae tweets' dialect led them to be labeled as less toxic \cite{sap2019risk}.
We conclude this study by discussing the limitations and ethical implications of the synthetic data, and cautioning against its real-world application.

Focusing on dialectal bias, our key assumption is that an \aae tweet and its corresponding \wae %
version should have the same toxicity label, therefore toxic \aae tweets whose \wae versions are non-toxic are candidates for label correction.%
\footnote{Note that this assumption does not hold for lexical items, because  substituting lexical items (e.g., swapping a minority mention for a majority mention)  would drastically change the denotational meaning of the sentence.}

However, gold-standard translations of \aae to \wae would require qualified translators, and automatic \aae-to-\wae translation systems do not exist, to the best of our knowledge. 
Therefore, we create a proof-of-concept study---we set up a \aae to \wae ``translation'' system using the few-shot capabilities of  the GPT-3 language model  \cite{brown2020language}.  
Under this mechanism, we prompt GPT-3 with four translation pairs \citep[taken from][]{Spears1998africanamerican} and an \aae tweet from our training data, and generate its \wae ``translation''.
The list of prompts, as well as further details, are provided in Appendix \ref{supp:gpt3}. Note that we do \emph{not} recommend this approach to build large scale parallel data for dialects, as discussed under ethical implications and limitations. %

Next, as per our heuristic, we only relabel toxic \aae tweets whose \wae translation is predicted as non-toxic by either our vanilla classifier trained  on the original \citet{founta2018} dataset, or an identical classifier trained on the \wae translated tweets.
The resulting dataset (\aae-relabeled) is the same size as the original dataset, but with 954 (12\%) out of 8260 toxic \aae tweets relabeled as non-toxic (examples in Table \ref{tab:qualitative_dialect}).
To assess the validity of the relabeling, the first three authors manually annotated toxicity of 50 randomly selected relabeled tweets.
On average, authors agreed with 84\% of the relabeling decisions.

Then, we evaluate the dialectal bias of \aae-relabeled and quantify the dialect and racial prediction biases from a RoBERTa-large classifier trained on \aae-relabeled, following \S\ref{sec:experiments_dialectal}.
As shown in the last row of Table \ref{tab:results_dialectal}, this relabeling scheme decreases dialectal bias more than any other debiasing method, specifically as measured by the FPR on \aae tweets, with one point drop in $F_1$ score. The $F_1$ score on the ``gold" test data (Table \ref{tab:results_dialectal}) are not fully reliable, as test data contain label biases and better performance could come from exploiting these biases. %
As shown in Table \ref{tab:userlevelrace18}, the model trained on \aae-relabeled has the lowest racial disparity in toxicity flagging rates compared to all other methods.

These results highlight that debiasing methods are much less effective at mitigating dialectal dataset biases compared to data relabeling.
For future investigations, we recommend obtaining human-written \aae-\wae pairs \cite[e.g., as done by][]{Groenwold2020aaeGPT2}. 
Additionally, to ensure less biased toxicity labeling, we recommend recruiting \aae speakers or experts for avoiding over-identification of \aae-markers as toxic \cite{Spears1998africanamerican,Croom2013-uc}.
Alternatively, we recommend exploring more holistic representations of social biases or toxicity %
\cite[e.g., Social Bias Frames;][]{sap2020socialbiasframes}.

\subsection*{Ethical Implications \& Limitations}
The above synthetic setting is meant to illustrate the role of labeling quality on biases in annotations.
We strongly caution against using this approach in real-world applications, such as building parallel datasets for dialects.
First, due to how its training data was selected, GPT-3 has likely not been exposed to many African American English varieties during training \cite{jo2020lessons}.
Second, pretrained language models are known to generate toxic language  at non-trivial rates \cite{Gehman2020RealToxicityPromptsEN}, which could cause differential toxicity in the translations.

\section{Related Work}
\label{sec:related}

\paragraph{Debiasing Toxicity Detection}
As the popularity of hate speech and toxic language detection systems has grown, several biases have been found in dataset and models, spurring various debiasing efforts to mitigate these individual biases \cite[e.g., gender bias, racial bias;][]{park-etal-2018-reducing,sap2019risk,davidson-etal-2019-racial}.
Some work tackles identity-based biases, e.g., using data re-balancing \cite{Dixon2018MeasuringAM}, or adversarial feature learning \cite{Vaidya2019EmpiricalAO}.
Less work has tackled racial or dialectal bias.
Notably, \citet{xia-etal-2020-demoting} use adversarial training to prevent the model from associating toxicity with \aae, showing only small improvements in fairness.
Based on those results, we do not explore adversarial methods, opting instead for ensemble-based methods of predefined bias reduction.
In contemporary work, \citet{10.1371/journal.pone.0237861} use a  re-weighting mechanism, which shows some effects in debiasing racial bias. We leave it for future work to evaluate this method in our setting. %
In contrast to all previous work, our experiments also measure the effectiveness of bias-agnostic methods.

\paragraph{Other General Debiasing Methods}
Several approaches for debiasing NLU tasks have been proposed lately.
Some approaches rely on adversarial training to remove protected attributes (e.g. gender or race), from a model’s internal representations \citep{zhang-2018-mitigating,Wang2019BalancedDA,xia-etal-2020-demoting}. 
Other approaches include confidence regularization \cite{utama-etal-2020-mind}, as well as other product of expert approaches \cite{he-etal-2019-unlearn,karimi-mahabadi-etal-2020-end} similar to the debiased training approach from \citet{clark-etal-2019-dont}, which is the only debiased training we employ due to its relatively strong performance.

\section{Conclusion}
\label{sec:discussion}
We investigate whether toxic language detection systems can be debiased using recently introduced methods for debiasing text classification in NLU tasks.
Focusing on two types of biases, lexical and dialectal, our experiments show that these methods face significant challenges in reducing the biased behavior in toxicity detectors.
This indicates that biases in toxic language detection might be different in nature compared to spurious associations studied in typical NLU settings.
We studied a synthetic scheme for relabeling examples with potential dialectal biases; our results indicate that correcting noisy labels results in better bias reduction.
Our findings suggest that instead of solely relying on development of automatic debiasing for existing, imperfect datasets, future work focus primarily on the quality of the underlying data for hate speech detection, such as accounting for speaker identity and dialect.
Indeed, such efforts could act as an important step towards making systems less discriminatory, and hence safe and usable.

\section*{Acknowledgments}
We thank the anonymous reviewers and Laura Vianna for helpful comments on this work.  This research was supported in part by NSF grants 1813153 and 1714566.

\bibliography{cleaned_ref}
\bibliographystyle{acl_natbib}

\clearpage
\appendix
\section*{Appendix}

\section{Further Details for Models}
\subsection{Model Debiasing}
\label{supp:predefined}

The \textsc{learned-mixin} ensemble allows the model to explicitly determine how much to trust the bias given the input:
\begin{align*}
\hat{p_i} =& \text{softmax} \{ \log(p_i) + g(\mathbf{x}_i) \log b_i \}
\end{align*} 
where $\mathbf{x}_i$ is the $i$th input text, $p_i$ and $b_i$ is the toxicity prediction produced by RoBERTa, and bias-only model respectively, and $g$ is a parametric function, which is defined as $\text{softplus}(\mathbf{w} \cdot \mathbf{h}_i)$, where $\mathbf{w}$ is a learned vector, $\mathbf{h}_i$ is the last hidden layer of the model for example $\mathbf{x}_i$, and the $\text{softplus}(x) = \text{log} (1+\exp{x})$. 
To prevent the \textsc{learned-mixin} ensemble from ignoring $b_i$, \citet{clark-etal-2019-dont} add an entropy penalty ($H$) to the loss:
\begin{align*}
R =& \alpha H(\text{softmax} \{ g(\mathbf{x}_i) \log b_i \}) 
\end{align*}
Where $H(z) = -\sum_j z_j \log z_j$
is the entropy and $\alpha$ is a hyperparameter.

\subsection{Data Filtering}
\label{supp:unspecified}
For the data filtering methods, we first filter data to 50\% of the original data as in \citet{swayamdipta2020datamaps}. Then we further downsample the dataset to 33\% of the original data to control that each training set has the same toxic ratio as the original training set. This step is to avoid confounding our results with different toxic ratio among different training sets.

\subsection{Training Settings}
\label{supp:training_setting}
For all the experiments, we fine-tune RoBERTa-large \cite{liu2019roberta} over the corresponding corpus with one GTX2080 Ti.
We use the default hyperparameters as provided in the \texttt{HuggingFace Transformers} library \citep{Wolf2019HuggingFacesTS}, with two major changes: we use a learning rate of $10^{-5}$ and 8 batch size in all experiments.

\subsection{Prediction Combining with Bias-only Model}
\label{supp:wbias_only}

To prevent the possibility that our \lmixin-\toxt/\oni is not well trained, thus resulting in the decrease of models' in-distribution performance, we use the joint-prediction from the main and bias-only model to infer the in-distribution test set and they obtain 94.15\% and 94.17\% accuracy, respectively. This is competitive performance as shown in Table \ref{tab:results_lexical}.

\section{Alternative Dataset of Toxic Language}
\label{supp:davidson-data}

\citet{davidson2017} collected data from Twitter, starting with 1,000 terms from HateBase (an online database of hate speech terms) as seeds, which the process relies on lexical biases. We find that performing data filtering methods over this dataset leads to degenerate behaviour. Specifically, as shown in Table \ref{apx:intrin_eval_word}, the easy region demonstrates least spurious correlation due to its heavily skewed class distribution, which further prevent us from downsampling to control the toxic ratio. We also train \lmixin-\toxt and \lmixin-dialect over the dataset. Table \ref{apx:word_bias_ex} shows that FPR of the debiased model increase instead except for the \oi category and Table \ref{apx:sty_eval}'s results behave in-line with Table \ref{tab:results_dialectal}.

\begin{table*}[t]
\centering
\small
\begin{tabular}{lccccc}
\toprule
 & Toxic Ratio &$R_{\noi} \downarrow$ & $R_{\oi} \downarrow$ & $R_{\oni} \downarrow$ & $R_{\aae} \downarrow$\\
\midrule
Original$\dagger$ & 0.8308 & 0.0287 & 0.4320 & 0.2610 & 0.4061\\
\midrule[0.03em]
Random  & 0.8312 & 0.0288 & 0.4312 & 0.2621 & 0.4011\\
AFLite  & 0.7669 & 0.0342 & 0.4708 & 0.2835 & 0.4236\\
DataMaps-Ambig. & 0.6736 & 0.0493 & 0.4683 & 0.3230 & 0.4445\\
DataMaps-Hard  &  0.6645 & 0.0521 & 0.4533 & 0.3190 & 0.4426\\
DataMaps-Easy & 0.9972 & 0.0135 & 0.0771 & 0.0396 & 0.0928 \\
\bottomrule
\end{tabular}
\caption{\label{apx:intrin_eval_word}
Lexical and dialectal associations between toxicity in the original dataset \cite{davidson2017} and various filtered counterparts. 
Random, AFLite, and DataMaps all contain only 50\% of the original data after filtering. (We could not perform downsampling on these datasets due to their heavily skewed label distribution.) 
Lower Pearson $R$ correlation value indicates less superficial patterns in the dataset, thus are less biased. The easy subset gives the best results here are due to its severe inbalanced label distribution.
}
\end{table*}

\begin{table*}[t]
\centering
\small
\begin{tabular}{lcccrccccc}
\toprule
 & \multicolumn{2}{c}{Test} & \multicolumn{2}{c}{\noi} & \multicolumn{2}{c}{\oi} & \multicolumn{2}{c}{\oni}\\
\cmidrule(lr){2-3} \cmidrule(lr){4-5} \cmidrule(lr){6-7} \cmidrule(lr){8-9}
 & Acc.$\uparrow$ & $F_1 \uparrow$ & $F_1 \uparrow$ & FPR$_{\noi} \downarrow $ & $F_1 \uparrow$ & FPR$_{\oi} \downarrow$ & $F_1 \uparrow$ & FPR$_{\oni} \downarrow$  \\
\cmidrule{2-9}
Original & 96.37 & 97.81 & 96.42 & 25.00 & 99.86 & 57.14 & 99.57 & 63.64 \\
\lmixin-\toxt & 96.15 & 97.69 & 96.19 & 28.57 & 99.78 & 42.86 & 99.28 & 72.73 \\
\bottomrule

\end{tabular}
\caption{\label{apx:word_bias_ex}
Lexical bias removal evaluation for debiasing methods. 
Original refers to the model trained over the full training set.
The test set is further categorized into tweets that contained relevant \toxt words. 
$F_1$ indicates models' performance while the false positive rate (FPR$_{\textbf{*}}$) reflects models' bias. 
The lower the FPR$_{\textbf{*}}$ is, the less biased the model tend to be.
}
\end{table*}

\begin{table}[!htbp]
\centering
\small
\begin{tabular}{@{}lcccc@{}}
\toprule
Debiasing Method &  & \multicolumn{3}{c}{Test}  \\
& $R_{\aae}$ & Acc. $\uparrow$ & $F_1 \uparrow$ & FPR$_\aae \downarrow$ \\
\midrule
Original & 0.4079 & 96.37 & 97.81 & 24.76    \\
\lmixin-Dialect & - & 96.48 & 97.88 & 22.86  \\
\bottomrule
\end{tabular}
\caption{\label{apx:sty_eval}
Dialectal bias evaluation for all debiasing methods, on both in-distribution test set as well as out-of-distribution dialect and race priming test sets.
In addition to accuracy and $F_1$, we report the false positive rate with respect to tweets in \aae (FPR$_{\aae}$), reflecting dialectal bias (lower is less debiased).
Each method is based on a RoBERTa-large classifier.
}
\end{table}

\section{Few-shot \aae-to-\wae Translation}
\label{supp:gpt3}

\textbf{Note that we do \emph{not} recommend the following approach to build large scale parallel data for dialects, as discussed under ethical implications and limitations (\S\ref{sec:gpt-3}).}

We use GPT-3 \cite{brown2020language} to create a few-shot \aae-to-\wae translation system, using the following set of example translation pairs drawn from \citet{Spears1998africanamerican}:

\begin{displayquote}
\aae: Get your triflin’ ass out of here. \\
\wae: Get your trifling self out of here.

\aae: I saw his ass yesterday.\\
\wae: I saw him yesterday. 

\aae: His ass is gonna get fried.\\ 
\wae: He is gonna get fried

\aae: Wassup, nigga?\\
\wae: What's up bro? 

\aae: $\langle$tweet$\rangle$\\
\wae:
\end{displayquote}

Note that \citet{Spears1998africanamerican} refers to \wae as White language varieties, and deals with English prevalent in the United States.

We prepend the formatted example pairs to each \aae tweet in our training data, and generate the translation from GPT-3 using top-0.95 nucleus sampling with a temperature of 0.5.
Prompts, formatting, and generation parameters were chosen based on manual inspection of the output.

\end{document}